\documentclass[conference]{IEEEtran}
\IEEEoverridecommandlockouts
\usepackage{float}
\usepackage{cite}
\usepackage{amsmath,amssymb,amsfonts}
\usepackage{graphicx}
\usepackage{textcomp}
\usepackage{xcolor}
\usepackage{algorithm}
\usepackage{algpseudocode}
\usepackage{subcaption} 
\usepackage{url}
\usepackage{array}
\usepackage{booktabs}  
\usepackage{listings}
\usepackage{xcolor}
\usepackage{tabularx}

\definecolor{background}{RGB}{255,255,255}
\definecolor{numb}{RGB}{125,125,125}

\lstdefinelanguage{JSON}{
    string=[s]{"}{"},
    stringstyle=\color{red},
    numbers=none,
    numberstyle=\small,
    stepnumber=1,
    numbersep=8pt,
    showstringspaces=false,
    breaklines=true,
    frame=lines,
    backgroundcolor=\color{background},
    literate=
     *{0}{{{\color{numb}0}}}{1}
      {1}{{{\color{numb}1}}}{1}
      {2}{{{\color{numb}2}}}{1}
      {3}{{{\color{numb}3}}}{1}
      {4}{{{\color{numb}4}}}{1}
      {5}{{{\color{numb}5}}}{1}
      {6}{{{\color{numb}6}}}{1}
      {7}{{{\color{numb}7}}}{1}
      {8}{{{\color{numb}8}}}{1}
      {9}{{{\color{numb}9}}}{1}
}

\def\BibTeX{{\rm B\kern-.05em{\sc i\kern-.025em b}\kern-.08em
    T\kern-.1667em\lower.7ex\hbox{E}\kern-.125emX}}
\begin{document}

\title{Efficient Multi-Hop Question Answering over Knowledge Graphs via LLM Planning and Embedding-Guided Search}

\author{
\IEEEauthorblockN{
Manil Shrestha, 
Edward Kim
}
\textit{ms5267@drexel.edu, ek826@drexel.edu}
\IEEEauthorblockA{
Department of Computer Science, Drexel University, Philadelphia, PA, USA 
}}

\maketitle

%

\begin{abstract}
Multi-hop question answering over knowledge graphs remains computationally challenging due to the combinatorial explosion of possible reasoning paths. Recent approaches rely on expensive Large Language Model (LLM) inference for both entity linking and path ranking, limiting their practical deployment. Additionally, LLM-generated answers often lack verifiable grounding in structured knowledge. We present two complementary hybrid algorithms that address both efficiency and verifiability: (1) LLM-Guided Planning that uses a single LLM call to predict relation sequences executed via breadth-first search, achieving near-perfect accuracy (micro-F1 $>$ 0.90) while ensuring all answers are grounded in the knowledge graph, and (2) Embedding-Guided Neural Search that eliminates LLM calls entirely by fusing text and graph embeddings through a lightweight 6.7M-parameter edge scorer, achieving over 100$\times$ speedup with competitive accuracy. Through knowledge distillation, we compress planning capability into a 4B-parameter model that matches large-model performance at zero API cost. Evaluation on MetaQA demonstrates that grounded reasoning consistently outperforms ungrounded generation, with structured planning proving more transferable than direct answer generation. Our results show that verifiable multi-hop reasoning does not require massive models at inference time, but rather the right architectural inductive biases combining symbolic structure with learned representations.
\end{abstract}


\section{Introduction}

Knowledge graphs (KGs) have emerged as powerful structures for representing domain-specific, structured information that supports verifiable, multi-hop reasoning. Meanwhile, large language models (LLMs) trained on vast web-scale corpora have achieved impressive fluency and generalization across a wide range of tasks. However, this same generality and end-to-end training on heterogeneous internet data introduce significant limitations.

Because LLMs are trained on broad, uncurated internet text, they do not always guarantee reliable factual accuracy in specialized or domain-specific contexts. Even when complete contextual information is supplied through prompting or retrieval augmentation, the results remain not fully reliable. LLMs are well known to produce hallucinations, i.e., outputs that are fluent and plausible yet factually incorrect or unsupported \cite{huang2025survey, dang2025survey, farquhar2024detecting}. Recent studies emphasize that hallucination remains one of the most pressing barriers to deploying LLMs in high-stakes scenarios.

In critical domains such as medicine, cybersecurity, and financial analysis, it is insufficient for an LLM to generate fluent answers; responses must be grounded in verifiable evidence, and the reasoning process should be transparent and traceable. Knowledge graphs (KGs) provide a strong foundation for such grounding by encoding structured domain knowledge and enabling explicit traversal and reasoning over relations~\cite{luo2024graph,yang2024integrated, wu2024kgv}. 
Recent work has also explored constructing KGs directly from unstructured text using LLM-based reasoning~\cite{kim2024structured}. Once such graphs are built, the ability to efficiently traverse them and answer natural language queries becomes crucial for practical deployment. Consequently, integrating KGs with LLMs has emerged as a promising direction for reducing hallucinations, improving factual consistency, and enhancing the interpretability of model outputs.

\begin{figure}[t]
    \centering
    \includegraphics[width=\linewidth]{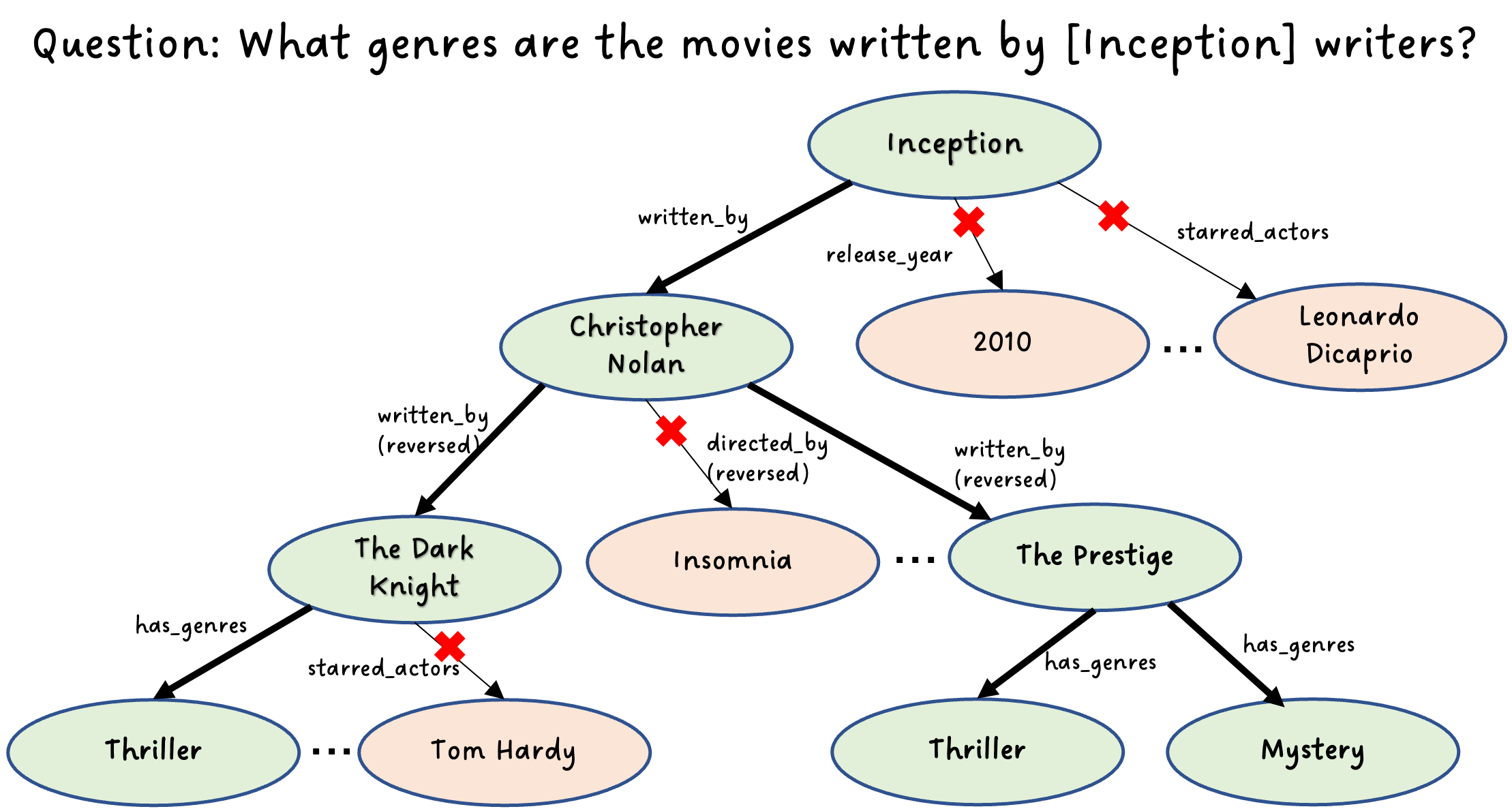}
    \caption{
Path pruning in knowledge graph exploration. Starting from the movie Inception, the goal is to find genres of other movies written by the same writers. Two strategies are discussed in this paper: (a) LLM-planned reasoning, where the model predicts the relation sequence 'written by → written by reversed → has genres'; and (b) multimodal edge scoring, which prunes paths using graph and text embeddings for efficient, LLM-free traversal.
}
    \label{fig:kg_exploration}
\end{figure}

Several recent methods, such as \textit{Plan-on-Graph} \cite{chen2024plan} and \textit{Think-on-Graph} \cite{sun2023think}, have explored using LLMs to plan reasoning paths directly over knowledge graphs. In these frameworks, the model performs iterative reasoning by generating traversal steps, one hop at a time, through repeated LLM invocations. While these methods improve interpretability and allow for natural language-based reasoning, they face serious scalability bottlenecks: each traversal step requires a separate LLM call, making the overall process computationally expensive, latency-prone, and often infeasible for large graphs or production-scale systems. As the number of nodes and relations grows, the combinatorial explosion in possible traversal sequences further amplifies these costs.

To address these challenges, we propose and experiment with two complementary strategies for efficient KG exploration. 
First, we investigate whether modern LLMs, with improved planning and reasoning capabilities, can identify the optimal sequence of relations to explore in a single or limited number of planning steps, thereby reducing repeated calls during multi-hop reasoning. 
Second, we explore a multimodal approach that combines text embeddings and graph embeddings to traverse the graph entirely without additional LLM invocations. 
This embedding-guided traversal provides a lightweight and scalable mechanism to approximate reasoning paths efficiently, offering significant improvements in speed and cost over when operating with large models that are expensive.

We argue that combining embedding-guided exploration with LLM planning offers a promising step toward \textit{trustworthy, efficient, and explainable} reasoning systems, paving the way for safer deployment of LLMs in high-stakes domains.

\section{Background}

\subsection{Knowledge Graphs and Multi-Hop Reasoning}
Knowledge graphs (KGs) represent structured world knowledge as collections of triples $(h, r, t)$, where $h$ and $t$ denote entities and $r$ represents the semantic relation between them \cite{ehrlinger2016towards, hogan2020knowledge}.
By capturing explicit relational information, KGs provide a foundation for interpretable reasoning across complex domains.
Reasoning over KGs can be performed through path traversal or logical inference, where multi-hop reasoning involves connecting distant entities via intermediate relations to answer complex queries \cite{bordes2013translating, sun2019rotate}.
For instance, to answer ``Which movies were directed by someone who also acted in \textit{The Terminal}?'' one must traverse multiple relations such as acted\_in and directed\_by.
Multi-hop reasoning thus enables compositional question answering and explainable inference by revealing the intermediate reasoning chain.

\subsection{Large Language Models and Hallucination Challenges}
Large language models (LLMs) such as GPT~\cite{brown2020language}, LLaMA~\cite{touvron2023llama}, and Qwen~\cite{yang2025qwen3} are pre-trained on massive text corpora drawn from the internet.
This broad pretraining grants them strong generalization and fluency but limits factual reliability in specialized or knowledge-intensive domains.
Since the underlying corpora are uncurated and they often exhibit hallucination,producing fluent but factually incorrect or unverifiable statements \cite{huang2025survey, dang2025survey, farquhar2024detecting}.
While retrieval-augmented generation (RAG) \cite{lewis2020retrieval} and prompting with contextual documents can improve factuality, such methods still rely on textual matching rather than structured reasoning.
As a result, LLMs struggle with logical consistency, relation chaining, and evidence traceability, especially in domains such as medicine, security, or finance where factual correctness is critical.

\subsection{Graph Embeddings and Semantic Search}

To enable efficient reasoning over large-scale graphs, numerous embedding-based approaches have been developed to encode entities and relations into continuous vector spaces.
Models such as TransE \cite{bordes2013translating}, DistMult \cite{yang2014embedding}, ComplEx \cite{trouillon2016complex}, and RotatE \cite{sun2019rotate} capture relational semantics, while random-walk methods like Node2Vec \cite{grover2016node2vec} and FastRP \cite{chen2019fast} exploit neighborhood co-occurrence structure.
These embeddings allow similarity-based retrieval and can guide reasoning by narrowing the search toward semantically relevant regions of the KG.
Embedding-based KGQA systems extend this idea by jointly modeling questions and graph elements in a shared space.
Methods such as EmbedKGQA \cite{saxena2020improving} and GraftNet \cite{sun2018open} align question embeddings with entities or relations to retrieve or traverse relevant subgraphs, effectively combining latent representations with structured reasoning.
Such techniques demonstrate how dense embeddings can serve both as retrieval cues and as heuristics for efficient multi-hop reasoning in large knowledge graphs.

\subsection{Grounded and Verifiable Reasoning}
Grounded reasoning aims to ensure that model-generated outputs are supported by verifiable evidence, rather than produced through unconstrained generation.
In the context of KG-based reasoning, grounding entails aligning each reasoning step with a corresponding graph edge or factual triple.
This enables transparent inspection of intermediate steps and supports external validation.
Verifiability is especially crucial in high-stakes domains, for instance, validating causal chains in biomedical research, verifying policy compliance in cybersecurity, or ensuring explainable audit trails in financial analysis.
Consequently, emerging research focuses on combining symbolic reasoning, graph embeddings, and LLM planning to achieve grounded, trustworthy, and cost-efficient reasoning pipelines.
Our work builds on this foundation by exploring efficient graph exploration mechanisms that preserve verifiability while reducing the computational overhead of LLM-driven reasoning.

\begin{figure*}[t]
    \centering
    \includegraphics[width=\linewidth]{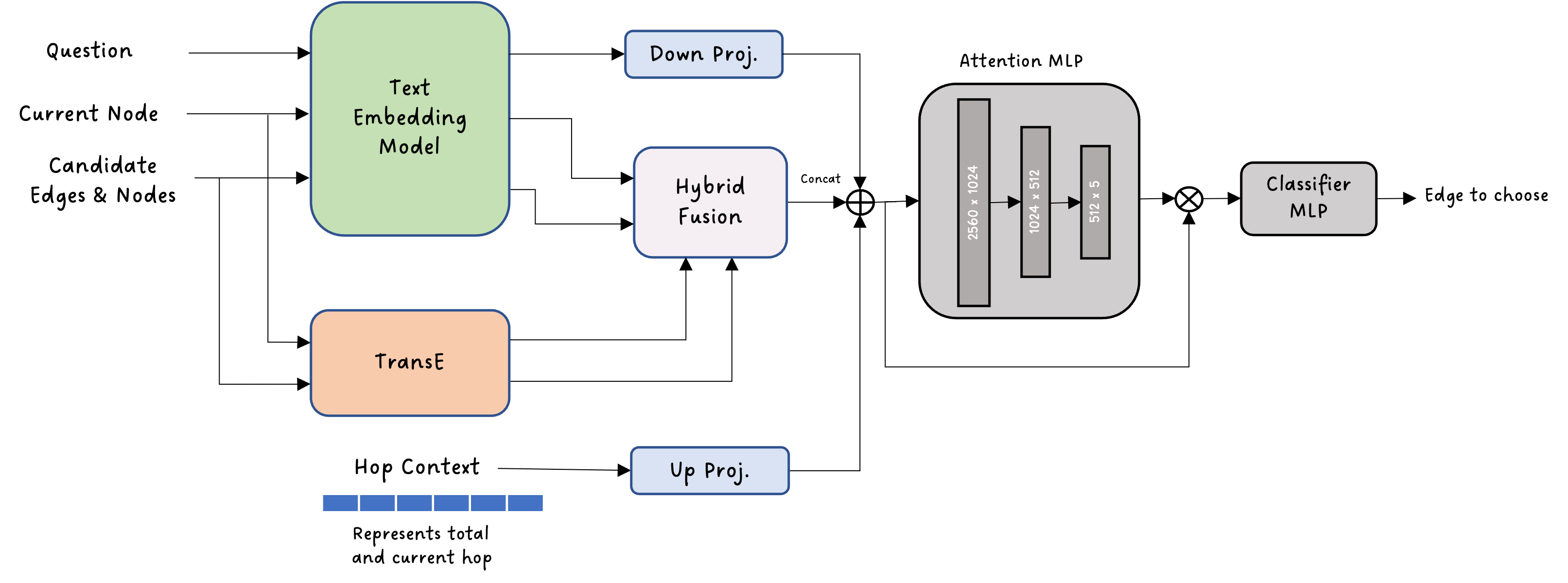}
    \caption{
Architecture of the proposed hybrid edge-scoring model.
The question text, current node, and candidate edges/nodes are encoded using both text and graph embedding models.
A hybrid fusion layer combines text and graph embeddings for each candidate triple.
The fused representations, together with the text embedding and hop context, are passed through an attention MLP to compute relevance weights.
The resulting attention-weighted features are fed into a classifier MLP, which outputs a score for each candidate edge.
}
    \label{fig:model_architecture}
\end{figure*}

\begin{figure}[t]
    \centering
    \includegraphics[width=\linewidth]{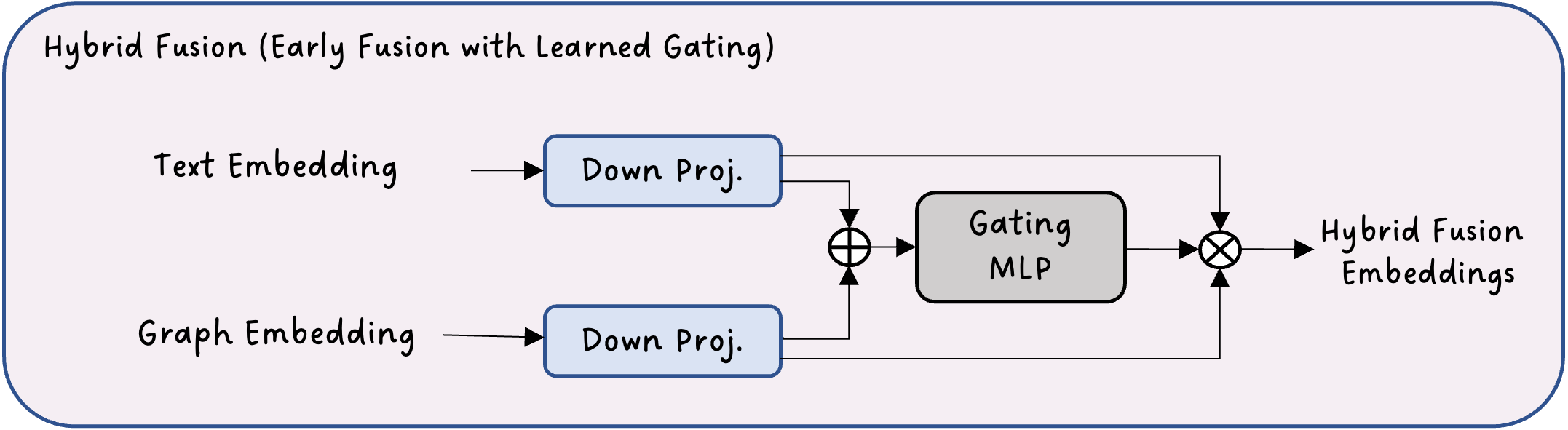}
    \caption{ \textbf{Hybrid Fusion}:
Text and graph embeddings are projected into a shared space and adaptively combined via a learned gating MLP to form hybrid embeddings that integrate semantic and structural information.
}

    \label{fig:hybrid_fusion}
\end{figure}

\section{Experiment Setup}
We evaluate our approaches on the MetaQA benchmark \cite{zhang2017variational}, a multi-hop question answering dataset over knowledge graphs.
The dataset is built on a movie knowledge base containing 43{,}234 entities, 134{,}741 triples, and 9 relation types.
Questions are grouped by reasoning complexity into 1-hop, 2-hop, and 3-hop categories, corresponding to single-step, two-step, and three-step reasoning over the graph.
MetaQA includes 407{,}513 natural language questions, with standard train, development, and test splits shown in Table~\ref{tab:metaqa_splits}.
We use the vanilla text version of the dataset with entity linking performed via exact string matching.

\begin{table}[h]
\centering
\caption{MetaQA dataset statistics.}
\label{tab:metaqa_splits}
\begin{tabular}{lccc}
\hline
\textbf{Split} & \textbf{1-hop} & \textbf{2-hop} & \textbf{3-hop} \\
\hline
Train & 96{,}106 & 118{,}980 & 114{,}196 \\
Dev   & 9{,}992  & 14{,}872  & 14{,}274 \\
Test  & 9{,}947  & 14{,}872  & 14{,}274 \\
\hline
\end{tabular}
\end{table}

\subsection{\textbf{LLM-Based Approaches for Knowledge Graph Question Answering}}
\label{subsec:llm_approaches}

We explore two complementary paradigms for applying large language models (LLMs) to multi-hop question answering over knowledge graphs. 
The first, a zero-shot baseline, directly queries an LLM without any symbolic grounding, while the second leverages the LLM’s reasoning ability to generate a structured traversal plan executed deterministically on the graph. 
Together, these variants capture the transition from purely neural reasoning to grounded neuro-symbolic reasoning.

We evaluate three language models for this experiment.
The primary model, GPT-5-mini (gpt-5-mini-2025-08-07), demonstrates strong reasoning ability and produces highly structured outputs with near-perfect planning accuracy. 
Qwen3-30B (qwen3-30b-a3b-instruct-2507-bf16)~\cite{qwen3technicalreport} serves as a self-hosted alternative based on a mixture-of-experts (MoE) architecture, offering competitive reasoning quality at zero inference cost. 
The 30B model is quantized to bfloat16 (bf16) precision to optimize GPU memory utilization without significant loss in output fidelity. 
Qwen3-4B (qwen3-4b-instruct-2507)~\cite{qwen3technicalreport} is a smaller, dense variant that enables faster inference with lower memory requirements, though its relation-planning consistency is comparatively lower than that of the 30B model.

All experiments were conducted using NVIDIA A40 GPU 48~GB VRAM. 
Model training and inference were managed through the vLLM framework~\cite{kwon2023efficient}, which enables efficient parallelized serving and optimized memory utilization. 
A single A40 GPU was used for model hosting and inference, while additional GPUs were allocated for fine-tuning and parallel evaluation. 
This setup allowed consistent comparison between API-based (GPT-5-mini) and self-hosted (Qwen3-30B, Qwen3-4B) models under the same hardware conditions.

\subsubsection{Zero-Shot LLM Baseline}
\label{subsubsec:zeroshot}

The zero-shot baseline quantifies how much factual and relational knowledge is encoded directly within the LLM parameters. In this setting, the model receives a natural language question and is instructed to produce all possible factual answers in a structured JSON format, returning up to 50 items per question. 

While models such as GPT-5-mini adhere closely to the structured format, the self-hosted Qwen models often produce malformed structured text. For these cases, a two-pass approach is adopted: the first pass generates free-form answers, and the second converts them into the required JSON schema. This separation reduces parsing failures from roughly 30–40\% to under 5\%, enabling consistent automatic evaluation against the ground-truth knowledge base.

This variant highlights the limitations of ungrounded LLM reasoning. Although the model can recall well-known facts about popular entities, it frequently returns the answers nonexistent in the graph. The results provide an upper bound on the LLM’s intrinsic factual knowledge and a lower bound on grounded accuracy achievable without knowledge graph access.

\subsubsection{LLM-Planned Relation Sequencing}
\label{subsubsec:llm_planned}
To address the shortcomings of ungrounded generation, the second variant introduces LLM-guided relation planning. 
Instead of producing final answers, the LLM predicts a sequence of relations that define a reasoning path through the knowledge graph. 
For example, given the question 
\textit{``What genres are films written by the writer of [Inception]?''}, 
the LLM outputs a structured plan:
\begin{center}
\texttt{\{"hops": [["written\_by"], ["has\_genre"]]\}}
\end{center}
This plan is then executed deterministically using a breadth-first search (BFS) over the knowledge graph, starting from the entity identified via bracketed mention (e.g., ``Inception'') and expanding only along the relations predicted for each hop.

This separation of reasoning and execution combines the semantic understanding of LLMs with the verifiability of symbolic reasoning. 
The LLM serves as a high-level planner, translating natural language into a relational program, while the BFS engine ensures that the resulting answers are grounded in the knowledge graph, eliminating possible hallucinations. The complete KG traversal logic is defined in Algorithm \ref{alg:llm-guided-bfs}. 
By exploring all entities reachable via the predicted relations, the approach achieves near-perfect accuracy on 1–2 hop questions and high performance even on 3-hop. 

The method offers several key advantages: (i) grounded correctness: every answer exists in the KB; (ii) interpretability: each reasoning step corresponds to an explicit relation; and (iii) flexibility: multiple relations per hop can handle ambiguous phrasing. 
However, the requirement of an LLM call per question adds cost and latency, motivating the next stage of our framework, where this reasoning ability is distilled into a smaller, self-hosted model.

\begin{algorithm}[t]
\caption{LLM-Planned Relation Sequencing}
\label{alg:llm-guided-bfs}
\begin{algorithmic}[1]
\Require Question $q$, KG $G=(V,E,R)$, LLM $\mathcal{M}$, hops $k$
\Ensure Answers $A \subseteq V$
\State $S \gets \textsc{ExtractEntities}(q)$
\State $H \gets \textsc{PlanRelations}(\mathcal{M}, q, R, k)$ \Comment{$H = [H_0,\dots,H_{k-1}]$}
\State $Q \gets \{(s,0): s\in S\}$,\; $A \gets \emptyset$,\; $Seen \gets \emptyset$
\While{$Q \neq \emptyset$}
  \State $(v,d) \gets Q.\textsc{dequeue}()$
  \If{$d = k$} $A \gets A \cup \{v\}$; \textbf{continue} \EndIf
  \If{$(v,d)\in Seen$} \textbf{continue} \EndIf
  \State $Seen \gets Seen \cup \{(v,d)\}$
  \ForAll{$u \in \bigcup_{r \in H_d} N_r(v)$}
    \If{$(u,d{+}1) \notin Seen$} $Q.\textsc{enqueue}((u,d{+}1))$ \EndIf
  \EndFor
\EndWhile
\State \Return $\textsc{Unique}(A)$

\end{algorithmic}
\end{algorithm}

\subsubsection{Knowledge-Distilled Relation Planner (LoRA)}
\label{subsubsec:kd_lora}

Results in Section~\ref{sec:discussion} show that the small Qwen3-4B model performs poorly at planning relation sequences, while GPT-5-mini achieves strong accuracy with consistent structured outputs. To reduce cost and remove API dependency without sacrificing grounding, we distill GPT-5-mini’s planning behavior into Qwen3-4B using supervised fine-tuning with Low-Rank Adaptation (LoRA)~\cite{hu2022lora}. The fine-tuning process is implemented using the LLaMA-Factory framework~\cite{zheng2024llamafactory}, which builds on HuggingFace Transformers and parameter efficient fine tuning (PEFT)~\cite{ding2023parameter} to provide efficient adapter-based training for instruction-tuned models. This setup allows the Qwen3-4B model to learn GPT-5-mini’s reasoning traces with minimal computational overhead, preserving structured output behavior while maintaining zero inference cost.

We first run the GPT-5-mini planner on a subset of the MetaQA training data (10{,}000 questions per hop level) to obtain teacher traces that include a brief reasoning string and a hop-by-hop relation sequence in JSON. These traces become instruction-tuning pairs for the student. We then fine-tune Qwen3-4B with LoRA (rank 16, $\alpha=32$, dropout 0.05, adapters on attention and feed-forward projections) so that the student learns to map questions to the same relation sequences. At inference, the fine-tuned model replaces GPT-5-mini in the planning stage, and the downstream graph traversal remains identical to Section~\ref{subsubsec:llm_planned}. As reported in Table~\ref{tab:accuracy_microf1}, this distillation substantially improves Qwen3-4B’s planning accuracy while keeping inference cost at zero and preserving full KG grounding and interpretability.

\subsection{\textbf{Embedding-Guided Neural Search}}
\label{subsubsec:neural_edge_scorer}

LLM-based relation planning, as discussed in Section~\ref{subsec:llm_approaches}, requires only a single model call to predict the full relation sequence, offering a significant efficiency advantage over multi-stage approaches such as \emph{Plan-on-Graph}, which invoke the LLM repeatedly across hops. 
To further reduce cost and latency, we introduce Embedding-Guided Neural Search, neural network model that eliminates even this single LLM call while preserving grounded reasoning. 
The key question is whether a lightweight neural model with only $\sim$6.7\,million parameters can approximate the relation-selection behavior of LLMs with hundreds of billions of parameters.

This approach replaces the LLM planner with a learned \emph{edge scorer} that predicts which outgoing edges from the current frontier should be traversed to answer multi-hop questions. 
The model fuses semantic information from text embeddings with structural information from knowledge-graph (KG) embeddings, and guides a deterministic beam-style breadth-first search (BFS) that expands only high-scoring relations and target nodes.

\subsubsection{Architecture}
The model architecture (Fig.~\ref{fig:model_architecture}) takes as input the question, current node, candidate edge, target node, and hop context. 
It comprises approximately 6.8M trainable parameters organized into modular components for representation, fusion, attention, and classification.

\paragraph{Input Representations.}
Each training instance includes five components: (1) a 1536-dimensional question embedding (OpenAI text-embedding-3-small)~\cite{openai_text_embedding_3_small}; (2--4) dual text--graph representations for the current node, candidate edge, and target node (each combining a 1536-dim text embedding with a 256-dim TransE embedding); and (5) a 6-dim hop-context one-hot vector encoding both the question’s required hop depth and the model’s current hop position. 
All components are projected into a shared 512-dimensional latent space.

\paragraph{Hybrid Fusion Module.}
For each graph component (node, edge, target), semantic and structural embeddings are combined through a Hybrid Fusion module (Fig.~\ref{fig:hybrid_fusion}). 
This module learns soft gating weights between text and graph signals, adaptively balancing semantic alignment and structural proximity. 
The fused outputs form modality-aware 512-dim representations $\mathbf{v}', \mathbf{r}', \mathbf{u}'$.

\paragraph{Cross-Component Attention.}
The five component embeddings $\{\mathbf{q}', \mathbf{v}', \mathbf{r}', \mathbf{u}', \mathbf{c}'\}$ are integrated via a \textbf{cross-component gated attention} mechanism. 
This layer learns dynamic weights over question, node, edge, target, and context features, allowing the model to emphasize different information sources depending on the reasoning stage (e.g., focusing more on the question in early hops and on target matching in later ones). 
This subnetwork accounts for roughly 46\% of total parameters ($\approx$3.1M).

\paragraph{Edge Scoring Classifier.}
The aggregated representation is passed through a three-layer MLP that outputs an edge-relevance score $s(q,v,r,u,c)\in[0,1]$, representing the likelihood that traversing $(v,r,u)$ lies on the correct reasoning path. 
The classifier uses ReLU activations, dropout (0.3), and a sigmoid output. 

Overall, the architecture efficiently integrates semantic and structural cues, enabling robust relation selection across reasoning hops while remaining lightweight enough for large-scale exploration.

\subsubsection{Training the classifier}
The edge scorer is trained end-to-end using a binary cross-entropy loss with class-imbalance weighting, where positive edges (those appearing on gold reasoning paths) are upweighted by the ratio of negative to positive examples in the training set. 
We use the AdamW optimizer~\cite{loshchilov2017decoupled} with learning rate $\eta = 10^{-4}$, weight decay $\lambda = 10^{-5}$, and gradient clipping at a maximum norm of 1.0 to stabilize training. 
The learning rate follows a cosine annealing schedule with warm restarts (T$_0$=10, T$_\text{mult}$=2, $\eta_\text{min}=10^{-6}$)~\cite{loshchilov2016sgdr}, which periodically resets the learning rate to escape local minima. 
We train for up to 50 epochs with a batch size of 1024 on an NVIDIA A40 GPU, employing early stopping (patience = 10). 
The model checkpoint with the highest validation F1 is retained for inference. 
Dropout (0.3) is applied to classifier layers during training to prevent overfitting, and all embeddings are passed through layer normalization before fusion to stabilize gradient flow across modalities.

\subsubsection{Inference via Neural Beam Search.}
During inference, all candidate edges from the current frontier are batch-scored on GPU. 
Edges are grouped by relation, averaged per relation, and the top-$B$ relations are selected for expansion, with each relation retaining the top-$M$ target nodes by score. 
Across $k$ hops, the algorithm maintains up to $O(B^k)$ partial paths but prunes them using multiplicative chain scoring,
\[
\text{ChainScore}(\rho) = \prod_{d=1}^{k} \bar{s}(r_d),
\]
which penalizes weak hops. 
In practice, we use $B\!\in\!\{3,5\}$ and $M\!\approx\!30$, balancing efficiency and coverage. 
Algorithm~\ref{alg:neural-beam-bfs} formalizes this procedure.

\begin{algorithm}[t]
\caption{Embedding-Guided Neural Edge-Scored BFS}
\label{alg:neural-beam-bfs}
\begin{algorithmic}[1]
\Require Question $q$, KG $G=(V,E,R)$, scorer $s(\cdot)$, hops $k$, beam $B$, cap $M$
\Ensure Answers $A\subseteq V$
\State $P \gets \{(\emptyset,S,[])\}$ \Comment{Paths: (relations, frontier, hop-context)}
\For{$d=1$ \textbf{to} $k$}
  \State $P'\gets\emptyset$
  \ForAll{$(\rho,F,\sigma)\in P$}
    \State $E_F\gets\{(v,r,u)\mid v\in F,(v,r,u)\in E\}$
    \State $\mathbf{s}\gets s(q,E_F,\text{hop}=d,\text{goal}=k)$
    \State Group $(v,r,u,s)$ by $r$; compute $\bar{s}(r)$ and select $\text{TopK}_r(\bar{s},B)$
    \ForAll{$r$ in top-k relations}
      \State Keep top-$M$ targets $u$ by score; update frontier $F'$ and hop-scores $\sigma'$
      \State $P'\gets P'\cup\{(\rho\!\cup\![r],F',\sigma')\}$
    \EndFor
  \EndFor
  \State $P\gets P'$ \Comment{Prune by partial chain score $\prod\sigma'$}
\EndFor
\State \Return $\textsc{Unique}\!\big(\bigcup_{(\rho,F,\sigma)\in P} F\big)$
\end{algorithmic}
\end{algorithm}

The neural scorer replaces costly LLM planning with a self-hosted, 6.7 M-parameter model ($\sim$14 MB, fp32) achieving millisecond-level batched scoring and $\sim$50–100 ms total per-question inference. 
It preserves grounded correctness (execution strictly within $G$), interpretability (explicit relation weights and gating behavior), and cost efficiency (no external API calls). 
Ablation 
studies removing graph embeddings show performance degradation 
(Appendix~\ref{app:ablation}), suggesting complementary roles of 
semantic and structural features.

\begin{table*}[ht]
\centering
\caption{Performance comparison across 1-, 2-, and 3-hop knowledge-graph question-answering (KGQA) tasks.
Results are averaged over all test questions within each hop depth. Micro F1 serves as the primary performance indicator, while Macro F1 provides class-balanced comparison. The table also reports efficiency statistics such as average nodes expanded and average inference time per query.}
\label{tab:accuracy_microf1}
\resizebox{\textwidth}{!}{
\begin{tabular}{lcccccccc}
\toprule
\textbf{Method} & \textbf{Hop} & \textbf{Hit Rate} & \textbf{Macro F1} & \textbf{Micro Precision} & \textbf{Micro Recall} & \textbf{Micro F1} & \textbf{Avg Time (s)} \\
\midrule
Zero-shot GPT-5-mini & 1-hop & 0.560 & 0.440 & 0.191 & 0.544 & 0.282 &  1.844 \\
                     & 2-hop & 0.247 & 0.229 & 0.272 & 0.299 & 0.285  & 1.965 \\
                     & 3-hop & 0.209 & 0.273 & 0.293 & 0.213 & 0.247 & 2.231 \\
\midrule
Zero-shot Qwen3-30B  & 1-hop & 0.320 & 0.281 & 0.132 & 0.326 & 0.188 &  0.503 \\
                     & 2-hop & 0.111 & 0.107 & 0.093 & 0.079 & 0.086  & 0.577 \\
                     & 3-hop & 0.111 & 0.170 & 0.204 & 0.084 & 0.119  & 0.531 \\
\midrule
LLM-Planned (GPT-5-mini) & 1-hop & 0.999 & 0.957 & 0.840 & 0.998 & 0.912 & 0.817 \\
                           & 2-hop & 0.999 & 0.993 & 0.967 & 1.000 & 0.983  & 1.213 \\
                           & 3-hop & 0.970 & 0.937 & 0.870 & 0.982 & 0.923 & 1.821 \\
\midrule
LLM-Planned (Qwen3-30B) & 1-hop & 0.962 & 0.935 & 0.843 & 0.952 & 0.894 & 1.045 \\
                          & 2-hop & 0.988 & 0.986 & 0.973 & 0.990 & 0.981  & 1.587 \\
                          & 3-hop & 0.495 & 0.467 & 0.849 & 0.469 & 0.604  & 1.715 \\
\midrule
LLM-Planned (Qwen3-4B) & 1-hop & 0.874 & 0.851 & 0.820 & 0.825 & 0.823  & 0.299 \\
                         & 2-hop & 0.752 & 0.742 & 0.890 & 0.526 & 0.661  & 0.205 \\
                         & 3-hop & 0.230 & 0.204 & 0.646 & 0.294 & 0.404 & 0.294 \\
\midrule
LLM-Planned (LoRA-Finetuned Qwen3-4B) & 1-hop & 0.999 & 0.996 & 0.837 & 0.997 & 0.910  & 0.490 \\
                         & 2-hop & 0.994 & 0.994 & 0.979 & 0.994 & 0.987  & 0.417 \\
                         & 3-hop & 0.863 & 0.843 & 0.927 & 0.884 & 0.905 & 0.547 \\
\midrule
Guided Neural Search (Path) & 1-hop & 0.996 & 0.973 & 0.937 & 0.968 & 0.952 & 0.229 \\
                                & 2-hop & 0.973 & 0.866 & 0.871 & 0.697 & 0.774  & 0.018 \\
                                & 3-hop & 0.911 & 0.673 & 0.728 & 0.446 & 0.553  & 0.588 \\
\midrule
Guided Neural Search (Greedy) & 1-hop & 0.996 & 0.975 & 0.936 & 0.984 & 0.959  & 0.007 \\
                             & 2-hop & 0.968 & 0.871 & 0.829 & 0.717 & 0.769 & 0.009 \\
                             & 3-hop & 0.952 & 0.729 & 0.769 & 0.562 & 0.649  & 0.016 \\
\bottomrule
\end{tabular}
}
\end{table*}

\begin{table*}[t]
\centering
\caption{Efficiency and cost comparison across methods on 3-hop KGQA tasks.}
\label{tab:efficiency_cost}
\resizebox{0.8\textwidth}{!}{
\begin{tabular}{lccc}
\toprule
\textbf{Method (3-hop)} & \textbf{Avg Nodes Expanded} & \textbf{Time (s)} & \textbf{Cost per Query (USD)} \\
\midrule
Zero-shot GPT-5-mini & 0.000 & 2.231 & 0.000698 \\
Zero-shot Qwen3-30B  & 0.000 & 0.531 & 0 (\textit{Self-hosted model}) \\
LLM Planning (GPT-5-mini) & 21.075 & 1.821 & 0.000356 \\
LLM Planning (Qwen3-30B) & 12.205 & 1.715 & 0 (\textit{Self-hosted model}) \\
LLM Planning (Qwen3-4B) & 9.239 & 0.294 & 0 (\textit{Self-hosted model}) \\
LoRA Finetuned Qwen3-4B & 20.034 & 0.547 & 0 (\textit{Self-hosted model}) \\
Guided Neural Search (Whole Path) & 65.523 & 0.588 & -- \\
Guided Neural Search (Greedy) & 14.884 & 0.016 & -- \\
\bottomrule
\end{tabular}
}
\end{table*}

\section{Discussion}
\label{sec:discussion}

Tables~\ref{tab:accuracy_microf1} and \ref{tab:efficiency_cost} summarize how grounding, planning, and embedding guidance shape the balance between accuracy, scalability, and cost in multi-hop knowledge-graph question answering (KGQA). Across all settings, the results indicate a consistent pattern: grounded reasoning outperforms memorized recall, and structured planning or learned scoring drastically improves efficiency without sacrificing verifiability.

\textbf{Understanding the Metrics:}
We report three complementary accuracy measures. 
\emph{Hit Rate} indicates whether at least one correct answer appears among the predictions, capturing coarse-level success. 
\emph{Micro-F1} aggregates precision and recall over all predictions, reflecting overall correctness weighted by relation frequency. 
\emph{Macro-F1} averages F1 across relation types, emphasizing balanced generalization across both common and rare reasoning patterns. 
Together, these metrics differentiate models that merely find a single correct answer (high Hit Rate) from those that consistently retrieve all correct ones with few false positives (high Micro/Macro-F1).

\subsection{Zero-Shot LLM Answers (Ungrounded Baseline)}
The zero-shot evaluation measures how much factual and relational knowledge can be recalled directly from model parameters without any grounding in the target knowledge graph. 
GPT-5-mini shows limited recall on 1-hop questions (micro-F1 = 0.28, hit rate = 0.56) and its accuracy declines further with reasoning depth (3-hop micro-F1 = 0.25). 
Notably, none of the correct answers appear within the model’s top-50 generated candidates for a large portion of the test questions. 
This suggests that while the model can produce semantically plausible entities, these guesses are not aligned with the actual contents of the knowledge graph.

Qwen3-30B, a smaller open-weight model, performs even worse in this ungrounded setting (3-hop micro-F1 = 0.12). 
Its generations are often fluent yet structurally invalid or incomplete, requiring a post-processing pass to enforce JSON formatting. 
Even when valid, the predictions tend to list plausible but nonexistent entities.

Overall, these results highlight a fundamental limitation of zero-shot language models when evaluated on unseen or proprietary graphs: 
their factual memory may contain fragments of world knowledge, but it cannot reconstruct relation chains or recover unseen entities without explicit grounding in the target graph. 
This underscores the need for hybrid approaches that constrain reasoning through structured graph traversal rather than relying solely on parametric memory.

\subsection{LLM-Planned Traversal (Grounded Reasoning)}
When we shift from free-form generation to LLM-planned traversal, accuracy increases dramatically. GPT-5-mini achieves nearly perfect performance on 1- and 2-hop tasks (micro-F1 ~ 0.91 – 0.98) and still sustains strong 3-hop performance (0.92). The reason is intuitive: instead of producing answers directly, the model predicts a structured relation plan that is later executed deterministically over the knowledge graph. Each reasoning step is thus verified against factual triples, eliminating hallucinations entirely.

Interestingly, the self-hosted Qwen3-30B also performs competitively under this paradigm (2-hop micro-F1 = 0.98), but its accuracy collapses at 3-hop (0.60). This sharp decline may stem from compounding errors in relation sequencing smaller decoding inconsistencies early in the plan propagate exponentially in multi-step traversal. The compact Qwen3-4B shows a similar pattern (1-hop = 0.82 → 3-hop = 0.40), confirming that while the planning mechanism is powerful, its success depends heavily on the model’s ability to produce stable-fully specified JSON structures and the ability to plan perfectly. Nevertheless, these grounded methods already outperform all zero-shot variants while maintaining full interpretability.

\subsection{Distilled Planner via LoRA (Small-Model Transfer)}
The LoRA-finetuned Qwen3-4B effectively bridges the performance gap between large and small models. Distilled on only about 10\% of the training data, it recovers most of GPT-5-mini’s planning accuracy achieving micro-F1 = 0.91 (1-hop), 0.99 (2-hop), and 0.91 (3-hop) while running locally at zero API cost. This striking improvement demonstrates that reasoning patterns from a large model can be compressed into a smaller student if the task is well-structured (here, predicting relation sequences rather than natural-language answers).

The practical implication is significant: grounded KGQA can now be executed entirely on a single commodity GPU. Cost per query drops to  zero (Table~\ref{tab:efficiency_cost}), and inference latency is reduced by more than twofold compared to API-based planning. Intuitively, LoRA distillation works well here because relation prediction lies at the intersection of symbolic and linguistic regularity, the mapping from question phrasing to relation type is highly compositional and therefore easy to transfer to smaller models.

\subsection{Embedding-Guided Neural Search (LLM-Free Inference)}
The embedding-guided models eliminate LLM calls altogether. Both the Whole-Path and Greedy variants use hybrid semantic–structural embeddings to score candidate edges and select the most promising traversal routes. Although trained on limited supervision, the models reach competitive accuracy: the Greedy variant attains micro-F1 = 0.96 (1-hop), 0.77 (2-hop), and 0.65 (3-hop) while completing each 3-hop query in only 16 ms - over 100 × faster than LLM-based planning (1.8 s for GPT-5-mini).

The performance drop with depth is intuitive. At higher hops, the model must integrate both long-range relational dependencies and semantic nuances, yet its scoring remains local to each edge. Errors accumulate multiplicatively when weak hops are chained. Still, the trade-off is favorable in high-throughput settings: these models can efficiently pre-prune the graph or serve as a low-cost first pass before invoking any LLM component.

\subsection{Interpreting the Trends}
Three trends emerge clearly:
\begin{enumerate}
    \item \textbf{Grounding converts recall into reasoning.} Once an LLM is forced to operate over KG structure rather than open-ended generation, factual accuracy rises dramatically because every hop is verifiable.
    \item \textbf{Planning generalizes better than generation.} Predicting relations instead of answers isolates the reasoning pattern from specific entity names, making it more transferable (as evidenced by LoRA distillation success).
    \item \textbf{Structure compensates for scale.} Even though Qwen3-4B has fewer parameters than GPT-5-mini, it reaches similar accuracy once provided the same structural bias showing that reasoning quality depends more on problem formulation than on raw scale.
\end{enumerate}

\subsection{Broader Implications and Future Work}
These experiments were conducted on MetaQA, a compact movie-domain KG with only nine unique relation types, but the insights extend to larger and more complex graphs. Most real-world knowledge graphs (e.g., Freebase~\cite{bollacker2008freebase}, UMLS~\cite{bodenreider2004unified}, or cybersecurity ontologies) also exhibit limited relation diversity relative to node count, making relation-centric reasoning approaches highly scalable.

Future directions include: 
(i) decomposing multi-hop questions into sequences of 1-hop subquestions that can be solved independently using a lightweight, fine-tuned LLM for single-hop QA transformation, followed by neural edge scoring for structured traversal; 
(ii) adapting the framework to noisy or incomplete entity-linking scenarios where graph coverage is partial; and 
(iii) extending evaluation to high-stakes domains such as medicine or finance, where verifiable multi-hop reasoning is essential. 

\subsection{Limitations}
While our methods demonstrate strong performance on MetaQA, 
several limitations warrant discussion. First, the evaluation 
is conducted on a single domain (movies) with exact entity 
linking; noisy or incomplete entity resolution may require 
additional robustness mechanisms. Second, MetaQA's nine relation types are relatively few compared to large-scale  knowledge graphs such as Freebase or Wikidata, which contain hundreds of relation types and billions of triples. Future 
work should validate scalability on such graphs and in domains requiring specialized reasoning, such as biomedical 
or legal knowledge bases.

\section{Conclusion}

We presented two complementary approaches for efficient and verifiable multi-hop reasoning over knowledge graphs that address the fundamental trade-off between accuracy, interpretability, and computational cost. Our LLM-guided planning method achieves near-perfect accuracy by separating high-level reasoning from deterministic graph traversal, ensuring all answers are grounded in verifiable triples while requiring only a single model invocation per query. Through knowledge distillation via LoRA fine-tuning, we further demonstrate that this planning capability can be compressed into a compact 4B-parameter model with minimal performance degradation, eliminating API dependencies and inference costs entirely.

Our embedding-guided neural search offers an alternative paradigm that removes LLM calls altogether, using a lightweight 6.7M-parameter model to fuse semantic and structural signals for edge scoring. While accuracy degrades slightly at deeper reasoning depths, this approach achieves over 100$\times$ speedup compared to LLM-based methods, making it practical for high-throughput scenarios or as a pruning mechanism in hybrid pipelines.

The findings suggest that effective knowledge graph question answering does not require massive language models at inference time. Rather, it requires the right architectural inductive biases that combine symbolic structure with learned representations. As knowledge graphs continue to scale in domains requiring trustworthy and auditable AI systems, such hybrid neuro-symbolic approaches provide a promising path forward for deploying verifiable reasoning in production environments.

\section*{Code and Data Availability}
The source code and trained models are publicly available at:

\noindent\url{https://github.com/ManilShrestha/big_data_workshop}

\bibliographystyle{unsrt}
\footnotesize
\bibliography{references}

\label{app:training}

\appendices

\section{Prompt for Hop-Wise Relation Selection}

\medskip
\noindent\textbf{Available relations:}
\textit{directed\_by, starred\_actors, written\_by, in\_language, has\_genre, release\_year, has\_tags, has\_imdb\_rating, has\_imdb\_votes}

\medskip
\noindent\textbf{Prompt Template (verbatim):}
\begin{verbatim}
You are analyzing a knowledge graph question 
to determine which relations to traverse.

Available relations in the knowledge graph:
{available_relations}

Question: "{question}"

This is a {max_hops}-hop question. You need to select 
which relation(s) might be relevant at EACH hop.

For each hop, select 1-2 most relevant relations from 
the list above. Think about the logical path needed 
to answer the question.

Examples:
- "Who directed movies starring [Actor]?" 
→ Hop 1: starred_actors, Hop 2: directed_by
- "What genre are films written by [Writer]?" 
→ Hop 1: written_by, Hop 2: has_genre
- "What year were movies by the director of [Movie]?" 
→ Hop 1: directed_by, Hop 2: release_year
- "What films can be described by [occupation]?" 
→ Hop 1: has_tags (films point TO tags/themes)
- "What films can be described by [Person Name]?" 
→ Hop 1: has_tags (films associated with person as tag)

Respond in JSON format:
{
  "reasoning": "brief explanation of the path",
  "hops": [
    ["relation1"],
    ["relation2"],
    ...
  ]
}

Provide your analysis:
\end{verbatim}

\section{Ablation Study}
\label{app:ablation}

Table~\ref{tab:ablation} presents ablation study results examining the contribution of each component in our edge scoring model. We evaluate four configurations: \begin{itemize} \item \textbf{TE+GE+HC} (Full Model): Combines text embeddings from a pre-trained language model (1536-dim), graph embeddings from TransE (256-dim), and hop context encoding. \item \textbf{TE+HC} (Abl1): Removes graph embeddings, relying only on semantic similarity from text representations and hop position. \item \textbf{TE} (Abl2): Uses only text embeddings, removing both graph structure and hop awareness. \item \textbf{GE+HC} (Abl3): Removes text embeddings, relying only on learned graph structure and hop position. \item \textbf{GE} (Abl4): Uses only graph embeddings, removing both textual semantics and hop context. \end{itemize} This design isolates the contribution of: (1) \textit{graph embeddings}, which capture structural knowledge from the knowledge graph; (2) \textit{text embeddings}, which provide semantic understanding of entities and relations; and (3) \textit{hop context}, which encodes the current position in the reasoning path.

\begin{table}[h]
\centering
\caption{Ablation study on embedding modalities. }
\label{tab:ablation}
\begin{tabular}{llcccc}
\toprule
\textbf{Config} & \textbf{Hop} & \textbf{Hit} & \textbf{Prec} & \textbf{Rec} & \textbf{F1} \\
\midrule
TE+GE+HC     & 1-hop & 0.996 & 0.937 & 0.968 & 0.952 \\
(Full Model) & 2-hop & 0.973 & 0.871 & 0.697 & 0.774 \\
             & 3-hop & 0.911 & 0.728 & 0.446 & 0.553 \\
\midrule
TE+HC        & 1-hop & 0.999 & 0.304 & 0.935 & 0.459 \\
(Abl1)       & 2-hop & 0.996 & 0.460 & 0.515 & 0.486 \\
             & 3-hop & 0.897 & 0.365 & 0.196 & 0.255 \\
\midrule
TE           & 1-hop & 1.000 & 0.304 & 0.930 & 0.458 \\
(Abl2)       & 2-hop & 0.994 & 0.451 & 0.504 & 0.476 \\
             & 3-hop & 0.898 & 0.371 & 0.198 & 0.258 \\
\midrule
GE+HC        & 1-hop & 0.948 & 0.278 & 0.843 & 0.418 \\
(Abl3)       & 2-hop & 0.881 & 0.265 & 0.297 & 0.280 \\
             & 3-hop & 0.461 & 0.114 & 0.060 & 0.079 \\
\midrule
GE           & 1-hop & 0.962 & 0.278 & 0.844 & 0.418 \\
(Abl4)       & 2-hop & 0.811 & 0.240 & 0.274 & 0.256 \\
             & 3-hop & 0.324 & 0.087 & 0.048 & 0.062 \\
\bottomrule
\end{tabular}
\end{table}

\textbf{Note:} The results reveal two key insights: (1) Text embeddings are the dominant modality, removing graph embeddings (Abl1, Abl2) causes minimal degradation compared to removing text embeddings (Abl3, Abl4), which severely impacts performance especially on multi-hop queries (3-hop F1 drops from 0.255 to 0.079). (2) Hop context provides marginal benefit when text embeddings are present (Abl1 vs Abl2), but offers more value with graph-only models (Abl3 vs Abl4, 3-hop F1: 0.079 vs 0.062). The full model's superior performance validates the complementary nature of combining both embedding modalities with hop awareness.
\begin{figure*}
  \centering
  \includegraphics[width=\textwidth]{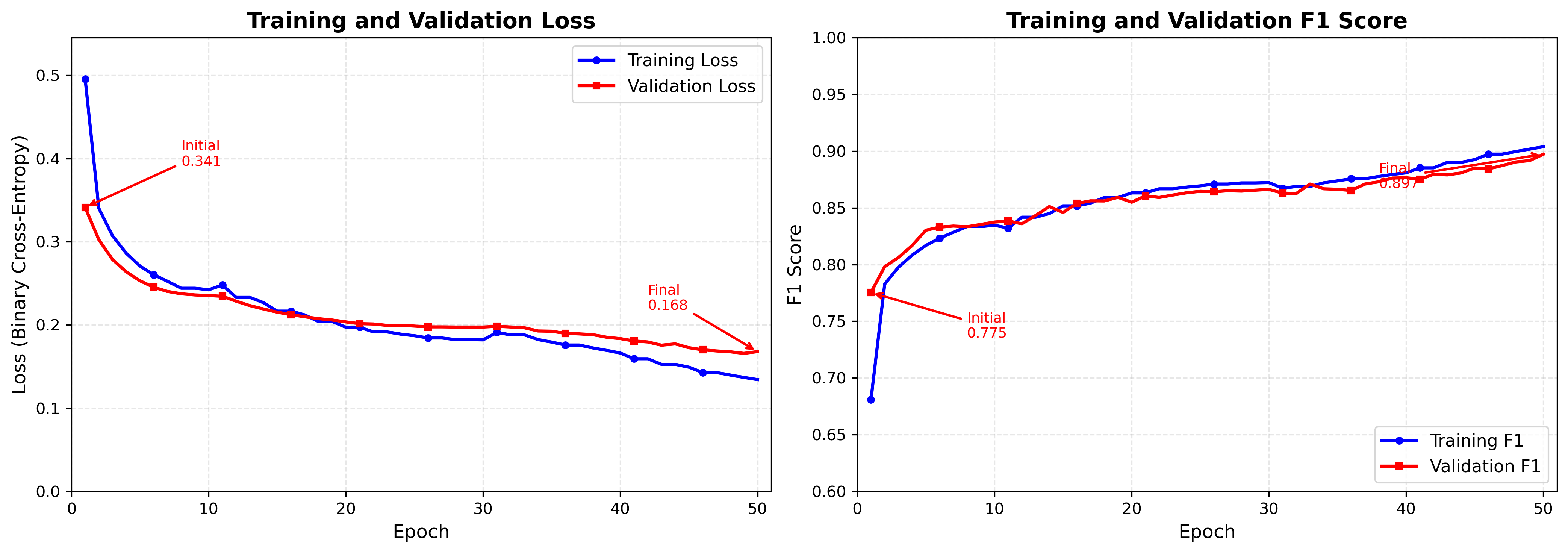}
  \caption{Training and validation loss for the Embedding-Guided Search model (Section~\ref{subsubsec:neural_edge_scorer}).}
  \label{fig:app-training}
\end{figure*}

\label{app:pareto}

\begin{figure*}
  \centering
  \includegraphics[width=\textwidth]{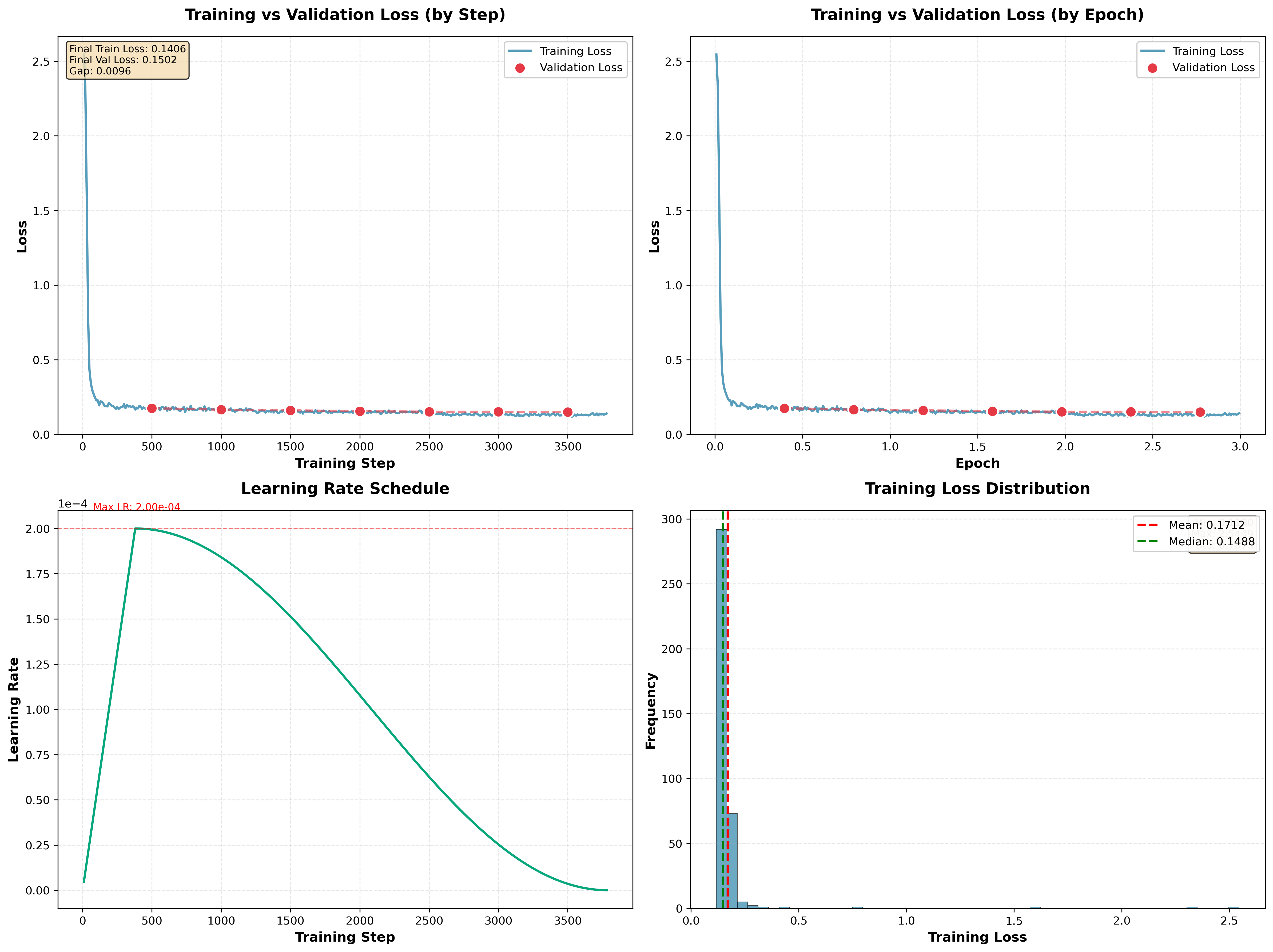}
  \caption{Training and validation loss for LoRA fine-tuning of the Qwen3-4B model using 10K training examples.}
  \label{fig:app-pareto}
\end{figure*}

\end{document}